\documentclass{bmvc2k}

\usepackage{amsmath}
\usepackage{mathtools}
\usepackage{amssymb}
\usepackage{array}
\usepackage{arydshln}
\usepackage{multirow}
\usepackage{booktabs}

\newcommand{\centered}[1]{\begin{tabular}{c} #1 \end{tabular}}

\title{Unsupervised Domain Adaptation by Uncertain Feature Alignment}

\addauthor{Tobias Ringwald}{tobias.ringwald@kit.edu}{1}
\addauthor{Rainer Stiefelhagen}{rainer.stiefelhagen@kit.edu}{1}

\addinstitution{
 Institute for Anthropomatics and Robotics (CV:HCI Lab)\\
 Karlsruhe Institute of Technology\\
 Karlsruhe, Germany
}

\runninghead{Ringwald et al.}{UDA by Uncertain Feature Alignment}

\def\eg{\emph{e.g}\bmvaOneDot}

\def\etal{\emph{et al}\bmvaOneDot}

\begin{document}

\maketitle

\begin{abstract}

Unsupervised domain adaptation (UDA) deals with the adaptation of models from a given source domain with labeled data to an unlabeled target domain. In this paper, we utilize the inherent prediction uncertainty of a model to accomplish the domain adaptation task. The uncertainty is measured by Monte-Carlo dropout and used for our proposed Uncertainty-based Filtering and Feature Alignment (UFAL) that combines an Uncertain Feature Loss (UFL) function and an Uncertainty-Based Filtering (UBF) approach for alignment of features in Euclidean space. Our method surpasses recently proposed architectures and achieves state-of-the-art results on multiple challenging datasets. Code is available on the project website.
\end{abstract}

\section{Introduction}

Training modern convolutional neural network (CNN) architectures with millions of parameters requires a vast amount of training data. However, data might be very expensive or difficult to acquire for a target domain while labeled data is readily available from another domain (source data). For example, the source domain could be constructed from synthetic data while classifying unannotated data (\eg medical images) is the actual inference task. 
Unfortunately, the domain difference between source and target data results in a severely degraded performance when evaluating a source-trained model on the new target domain. 

Unsupervised domain adaptation seeks to address this domain shift problem in order to maximize a model's accuracy on unlabeled target images given only labeled data from a source domain. Several different approaches have been proposed in recent research to achieve this goal. Pixel-level methods try to manipulate source and target images through style-transfer by mapping them into a joint image space so that a common classifier can be used. At feature level, methods either try to minimize distribution divergence measures such as the maximum mean discrepancy (MMD)~\cite{can}, Kullback-Leibler divergence~\cite{meng2018adversarial} or reach feature similarity by enforcing domain confusion between source and target features through adversarial training~\cite{simnet,hoffman2017cycada}. Other approaches -- such as \cite{long2018conditional,han2019unsupervised,manders2018adversarial} -- have combined this with a model's predictive uncertainty by \eg forcing the uncertainty distribution to be similar on source and target data. In this paper, we also explore the usage of uncertainty for unsupervised domain adaptation, which we quantify under the Monte Carlo dropout~\cite{mcdropout} approximation of Bayesian inference. However, in contrast to prior work, we leverage the uncertainty for feature alignment in Euclidean space and also for filtering of target data instances. 
Furthermore, we identify batch-normalization~\cite{batchnorm,dsbn} as point of failure for domain adaptation training on multiple GPUs and present a concept based on ghost batch-normalization~\cite{ghostbatchnorm} to fix this problem.
Our thorough experimental section shows that our proposed approach achieves state-of-the-art results on popular UDA benchmark datasets.

To summarize, our contributions are as follows: 
(i) We propose a new loss function that exploits a model's uncertainty for feature alignment in Euclidean space and filtering of uncertain pseudo-labels. 
(ii) We extend the concept of ghost batch-normalization to UDA setups and propose the smart batch layout (SBL). 
(iii) In combination, our proposed approach achieves state-of-the-art (SOTA) results on multiple benchmark datasets such as Office-Home~\cite{officehome} and Office-Caltech~\cite{officecaltech}. Code will be made available to the community to encourage the reproduction of our results.

\begin{figure}[t!]
\vspace{0.5cm}
\begin{tabular}{cc}
\centered{\bmvaHangBox{{\includegraphics[width=5cm]{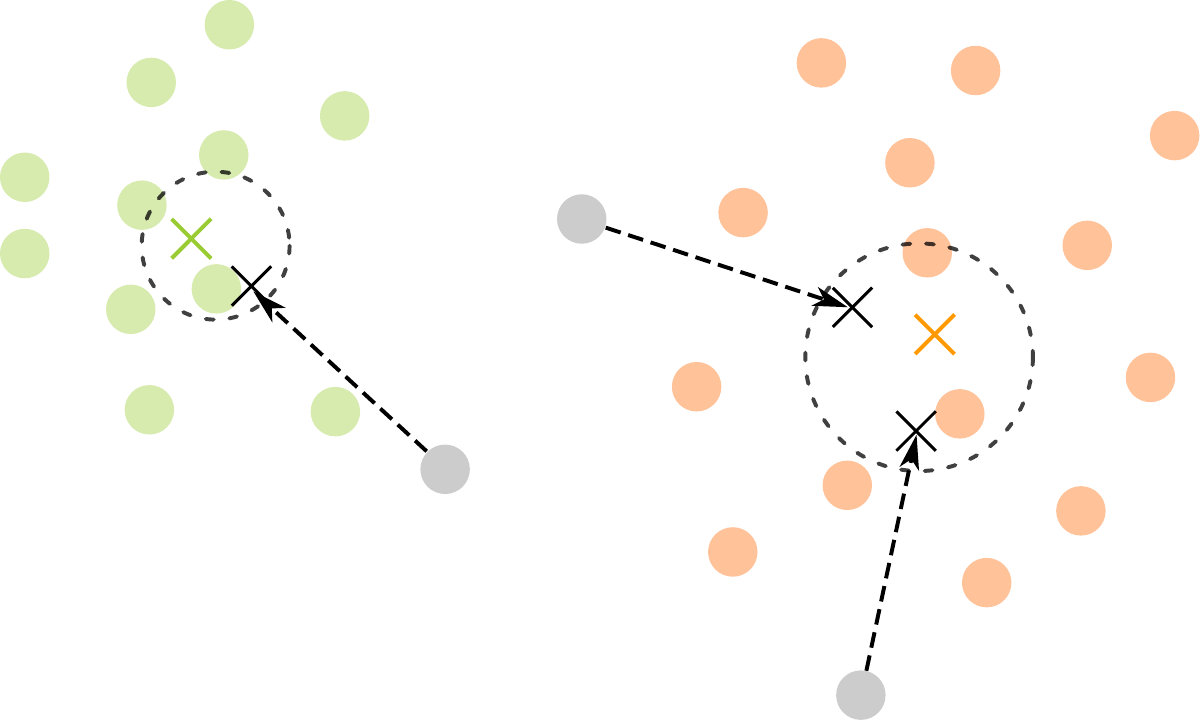}}}}&
\centered{\bmvaHangBox{{\includegraphics[width=6cm]{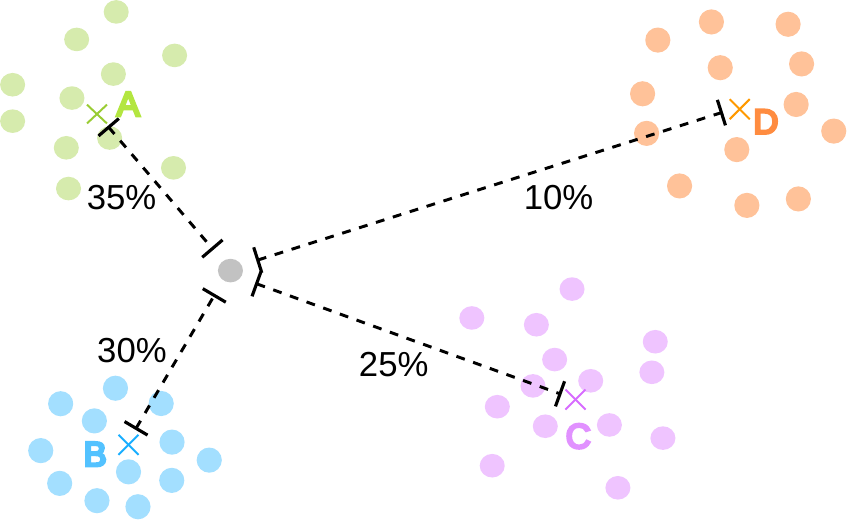}}}}\\
(a)&(b)
\end{tabular}
\caption{Our two applications of uncertainty: (a)~Iteratively reassigned Uncertain Feature Means (UFM). Instead of having a fixed position, feature means move depending on the assignment of grey samples. In a 2D example, we can visualize the uncertain area with dashed circles. In higher dimensions, this produces a hypersphere. (b)~Uncertainty-based distances for the Uncertain Feature Loss (UFL). The distance of the current sample (grey blob) is adjusted based on its uncertain assignment to the class set $\{A, B, C, D\}$.}
\label{fig:teaser}
\end{figure}

\section{Related Work}

In prior research, the domain shift problem was tackled in multiple different ways: The authors of \cite{atapour2018real} propose an image-level domain adaptation approach that leverages style transfer. Synthetic data is used for training and then transferred into the real domain to achieve domain adaptation for monocular depth estimation. Similarly, Bousmalis~\etal \cite{bousmalis2017unsupervised} also use a GAN-based approach to modify source examples and make them appear as if drawn from the target domain. Methods based on feature-level adaptation have also been proposed:
One of the first works in this direction was proposed by Ganin~\etal~\cite{ganin2014unsupervised}. They present a gradient reversal layer that is attached to a feature extractor. This often called RevGrad layer forces the feature distributions from the source and target domain to be as indistinguishable as possible, thus yielding domain-invariant representations. 
Similarly, Pinheiro~\etal~\cite{simnet} use a prototype-based algorithm that forces features extracted by a CNN to be domain-invariant. They propose a setup that eventually learns a pairwise similarity between said prototypes and images from the target domain. 
Another popular way to match feature representations between domains is based on maximum mean discrepancy. While already used by Long~\etal~\cite{long2013transfer}, it was recently picked up again by Kang~\etal~\cite{can} and achieves domain adaptation by minimizing this measure between the source and target distributions. Chang~\etal~\cite{dsbn} follow another direction and propose domain-specific batch-normalization layers to capture the different distributions between the domains.
Other approaches combine these feature- and image-level approaches. Hoffman~\etal~\cite{hoffman2017cycada}, for example, proposed the CyCADA framework, that adapts feature representations at both the feature- and pixel-level by enforcing local and global structural consistency. Additionally, they use cycle-consistent pixel transformation, which they show to be important for their semantic segmentation task.

The usage of a prediction model's uncertainty was also the subject of prior research in domain adaptation. The foundation for many works using uncertainty is the popular Monte Carlo dropout (MC dropout)~\cite{mcdropout}. MC dropout leverages the standard dropout layer~\cite{dropout} during inference time to get varying probability outputs which can be seen as an approximation to Bayesian inference in deep Gaussian processes. Long~\etal~\cite{long2018conditional}, for example, control the classifier uncertainty to guarantee the transferability and also condition a domain discriminator on the uncertainty of classifier predictions. Han~\etal~\cite{han2019unsupervised} propose the calibration of predictive uncertainty of target domain samples given the source domain uncertainties. They quantify their model's uncertainty by a Bayesian Neural Network under a general R\'{e}nyi entropy regularization framework.
Manders~\etal~\cite{manders2018adversarial} make use of an adversarial approach that enforces the target domain uncertainties to be indistinguishable from the source domain uncertainties.
In this work, we explore the use of uncertainty in a different way than prior work: We do not use adversarial or pixel-level methods but instead consider uncertainty as a distance measure at the feature level. Given a CNN feature extractor and uncertainties obtained through MC dropout, we align the features in Euclidean space in a way that reflects the model's uncertainty. This helps to separate distinct classes during early training while keeping together instances with high confusion potential until their certainty level rises. In a similar way, we exploit a model's uncertainty to filter low quality pseudo-labels and defer their usage for later training stages.

\section{Approach}

In an unsupervised domain adaptation (UDA) setup, we consider a labeled source domain dataset $\mathcal{D}_s=\{( x_s^{(i)}, y_s^{(i)}) \}_{i=1}^{N_s}$ and unlabeled target domain dataset $\mathcal{D}_t=\{x_t^{(i)}\}_{i=1}^{N_t}$, where $x_s^{(i)}$ is the i-th example in the source domain and $y_s^{(i)} \in \mathcal{C}$ is its corresponding label from label set $\mathcal{C}$ with $\lvert \mathcal{C} \rvert = N$ classes. The objective is then to predict the associated ground truth label $y_t^{(i)}$ for a given $x_t^{(i)}$.
For our approach, we utilize a common CNN feature extractor ($f(\cdot)$, such as ResNet~\cite{he2016deep}) in conjunction with a classifier ($cl(\cdot)$) that predicts class probabilities.
We denote their combination as $g(\cdot)$ with trainable weights $\theta$.

With this setup, we can now formally introduce our concept of uncertainty using MC dropout. Let $N_\text{MC}$ be the number of stochastic forward passes using MC dropout. We can then calculate the new class probabilities as 
\begin{equation}
    \Tilde{p}(x_t^{(i)}) = \frac{1}{N_\text{MC}} \sum_{j=1}^{N_\text{MC}} cl\left( f(x_t^{(i)}) \odot m_j\right), \label{eq:mc_dropout}
\end{equation}
where $m_j$ is a mask drawn from a Bernoulli distribution according to the dropout rate and~$\odot$~is the element-wise multiplication. We interpret the averaged probabilities $\Tilde{p}(\cdot)$ as a proxy measure for how uncertain (or certain) the model is in its predictions. Under this definition, a model would be completely uncertain of its prediction when $\Tilde{p}$ is equal to the uniform distribution and completely certain when it can be expressed by the Kronecker delta function. Usually, the probability distribution exhibits a few distinct peaks. These are the classes the model is confused by and which it can not separate given the current training progress. For brevity, we will use $p$ to denote the usual softmax probabilities of a network and $\Tilde{p}$ to denote the probability distribution obtained by averaging $N_\text{MC}$ MC dropout iterations.
In the following sections, we will describe how this uncertainty measure is used to achieve UDA.

\subsection{Binned Instance Sampling}

As the foundation of our method, we propose a new sampling scheme called Binned Instance Sampling (BIS). Before applying BIS, we first train our classification network $g$ with weights $\theta$ in a supervised fashion on data tuples $(x_s^{(i)}, y_s^{(i)}) \in \mathcal{D}_s$. This already enables us to obtain a probability distribution 
$p_t^{(i)}= \{ g_\theta(x_t^{(i)})  | x_t^{(i)} \in \mathcal{D}_t \} $ for every target domain image. Every few training steps, we estimate pseudo-labels as $\hat{p}_t^{(i)}=\underset{c \in \mathcal{C}}{\text{argmax}}\: p_{t,c}^{(i)}$ and associate it with the i-th example in the target domain. Based on this, we start the actual domain adaptation phase. 
Evidently, a model only trained on the source domain will suffer from the domain shift problem and produce noisy label estimates. The goal of BIS is to reduce the impact of wrong pseudo-labels by preferring high confidence examples while still considering the whole target domain dataset for training. 

Given $N_B$ bins of decreasing size $\lambda=\{\lambda_0, ..., \lambda_{N_B-1}\}$, BIS first groups target domain examples into structure $\kappa_{1..\lvert C \rvert}$ based on their pseudo-label assignment and then sorts them in descending order based on their softmax probability $p_{\hat{p}}$ where $\hat{p}$ was obtained earlier by pseudo-labeling. At this point, a class $c \in \mathcal{C}$ is drawn. We then randomly sample instances from $\kappa$ according to Equation~\ref{eq:sampling}.
Here, $\text{sample(a, b)}$ randomly takes $b$ samples from the list of samples $a$ and $[\cdot, \cdot]$ is the slicing operator.
\begin{equation}
    \mathcal{T}_{c} = \bigcup_{i=0}^{N_B-1} \text{sample}\left(\kappa_c\left[i\, \frac{\lvert \kappa_c\rvert}{N_B}, (i+1)\frac{\lvert \kappa_c\rvert}{N_B} \right], \lambda_i\right) \label{eq:sampling}
\end{equation}
For a batch $\mathcal{B}$ with size $\lvert \mathcal{B} \rvert$, this sampling process is repeated until $\lvert\bigcup_c \mathcal{T}_{c}\rvert = \frac{\lvert \mathcal{B \rvert}}{2}$ is reached where $\mathcal{T}_c$ is the set consisting of $\sum_j \lambda_j$ instances of randomly sampled class $c$. The other half of the batch is similarly sampled from the source domain and denoted by $\mathcal{S}$. In this case, however, the order of samples is irrelevant. The final result of BIS is two lists $\mathcal{S}$ and $\mathcal{T}$ both containing instances for the same classes. Note that $\mathcal{T}$ is based on fuzzy pseudo-labels and might contain instances that do not correspond to the sampled class label. The further usage of these two lists for training is discussed in the following section.

\subsection{Smart Batch Layout}\label{sec:batchnorm}

The batch-normalization layer first proposed by Ioffe~\etal~\cite{batchnorm} is an important part of most modern neural network architectures. Batch-normalization is said to address the internal covariate shift problem and enables faster training with higher learning rates. During training, batch-normalization whitens a given input feature map $x$ by applying Equation~\ref{eq:batchnorm}. Here, $\mu$ and $\sigma^2$ are the channel-wise mean and variance of all channel-slices in a mini-batch. $\gamma$ and $\beta$ are learnable parameters that can \textit{scale} and \textit{shift} the outputs if necessary. Additionally, an exponential moving average is calculated over the mini-batch $\mu$ and $\sigma^2$ values while training. During test time, these aggregated population statistics are used instead of the current mini-batch statistics.
\begin{equation}
    \mathrm{BatchNorm}(x)\,=\,\gamma \frac{x-\mu}{\sqrt{\sigma^2 + \epsilon}} + \beta\label{eq:batchnorm}
\end{equation}

In modern deep learning frameworks (\eg PyTorch~\cite{paszke2017automatic}) models can often be distributed over multiple GPUs. To avoid synchronization overhead during training, these frameworks calculate the mean and variance only over the partial mini-batch on the local replica, \eg for a batch size of 128 and 4 GPUs, every replica would consider only 32 examples. Updates to the population statistics are only done on the first replica and then broadcasted before the next forward pass. While this setup -- sometimes referred to as ghost batch-normalization -- is not the correct implementation w.r.t.~\cite{batchnorm}, it can actually help to improve results~\cite{ghostbatchnorm}. 

\begin{figure}[t!]
\vspace{0.25cm}
\centering
\includegraphics[width=0.9\textwidth]{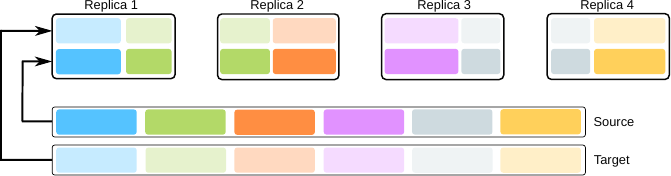}
\caption{Visualization of our proposed smart batch layout (SBL) with 4 replicas, 6 classes and 2 domains. Each colored box is representing multiple images of a certain class.
}
\label{fig:smart_batch_layout}
\end{figure}

However, this can be problematic in UDA setups. As per definition of the domain adaptation problem, the source and target data are drawn from a different distribution. Consider \eg a source domain of synthetic data (gray scale images) and a target domain of real world images (full RGB): When using multiple GPUs for training, the updates to $\mu$ and $\sigma^2$ are now highly dependent on the order of examples within a mini-batch as the statistics are only computed per replica. If exclusively source instances are on the first replica, only their means and variances will contribute to the test time population statistics; the same holds true for other extreme constellations. This hinders domain adaptation performance when using the same network as pseudo-labeler, as the statistics are now adjusted for the source distribution. We propose the smart batch layout to counteract this problem: Given the two lists $\mathcal{S}$ and $\mathcal{T}$ from the previous section and number of replicas $N_R$, we distribute the sampled instances so that every replica contains $\frac{\lvert \mathcal{B} \rvert}{2\times N_R}$ samples from both $\mathcal{S}$ and $\mathcal{T}$. Examples from a single class ($\mathcal{S}_c$ and $\mathcal{T}_c$) are first equally exhausted before considering the next class. Overall, this evenly distributes classes and domains so that every replica considers both the source and target distributions for local statistics computations. The same concept applies to the first replica responsible for the aggregation of population statistics and is visualized in Figure~\ref{fig:smart_batch_layout}.

\subsection{Uncertain Feature Loss and Filtering}

Only training on pseudo-labels generated by our model $g$ leads to degraded performance, because wrong predictions are eventually considered as training targets during the following iterations. We thus try to make the model aware of its own prediction uncertainty and consider other plausible classes instead of only the maximum prediction. Recall from above our basic training loop: After training $g$ on the source domain, we start the adaptation phase by generating pseudo-labels ($\hat{p}^{(i)}$ from distribution $p^{(i)}$) every few mini-batches. Additionally, we calculate the uncertainty distribution $\Tilde{p}$ and features $\Upsilon = \{\forall i: \upsilon^{(i)} = f(x^{(i)})\}$. 
We then generate mean features for all classes. However, we do not rely on the maximum prediction $\hat{p}$ but instead assign a class based on the uncertainty $\Tilde{p}$ by conducting weighted random sampling $\overline{p}^{(i)} = \text{weightedSampling}(\mathcal{C}, \Tilde{p}^{(i)})$. Given these assignments, we then construct uncertain mean features (UFM) for every class as $\Omega(c) = \frac{1}{\lvert \iota_c \rvert} \sum \iota_c  $, where $\iota_c = \{\upsilon^{(i)} | \upsilon^{(i)} \in \Upsilon \land \overline{p}^{(i)} = c\}$. 
Because $\overline{p}$ is resampled every few steps, this does not result in a fixed feature mean, but instead in an uncertain hypersphere around the class mean embedding based on the current iteration's assignments (visualized in Figure~\ref{fig:teaser}a). 

The distribution $\Tilde{p}_c^{(i)}$ can be interpreted as a measure for how likely the model currently thinks the i-th example belongs to class $c$. We propose to use this information as a distance measure in order to align the features in Euclidean space according to the uncertainty. Intuitively, the distance between a feature embedding and a class mean embedding should be minimized when the model is certain about its assignment and maximized when it is uncertain. Figure~\ref{fig:teaser}b visualizes this in a toy example. There, the current sample $x^{(i)}$ (grey disk) has $\Tilde{p}^{(i)}=\{0.35, 0.30, 0.25, 0.10\}$ for the 4 classes. We now want to align $f(x^{(i)})$ in a way that its distance to the uncertain class means reflects $\Tilde{p}^{(i)}$: Distances to more likely assignments are kept low, while unlikely ones are pushed further away in comparison. As a result, highly unlikely classes (such as D) are separated from $x^{(i)}$ due to their low probability so that the model only needs to discriminate a reduced subset of possible assignments during following training steps. This can be seen as a deferred feature disentanglement: First, easy to distinguish classes are separated from each other due to having a single peak in the uncertainty distribution, thus minimizing the distance towards that class. As training progresses, the model sees more and more target data which improves its feature estimation and prediction qualities. For hard to classify examples, this shifts the multimodal uncertainty distribution towards a Kronecker delta, eventually forcing the model to also separate the more difficult samples. Implementation-wise, we achieve this by enforcing the softmax-normalized negative $\ell_2$ distances to all class mean embeddings to reflect $\Tilde{p}^{(i)}$.
With these preconditions, we can write our proposed Uncertain Feature Loss (UFL) loss function as a combination of above uncertain feature alignment (distance-based) and the current pseudo-label $\hat{p}$ (prediction-based) in Equation~\ref{eq:ufl}.
\begin{align}
\mathcal{L}_\text{UFL}(x,\, \Omega) &= -\sum_{c \in \mathcal{C}} \overbrace{\Tilde{p}_c\, \mathrm{log}\: \zeta^2\left(\left\{\omega \in \Omega: \text{-}\lvert\lvert f_\theta(x) - \omega \rvert\rvert_2^2 \right\} \right)_c}^{\text{distance-based}}  - \sum_{c \in \mathcal{C}} \overbrace{\phi(\hat{p})_c\: \mathrm{log}\left( g_\theta(x)_c \right)}^\text{prediction-based} \label{eq:ufl}\\  \text{where}\ \ \zeta(z)_k &= \frac{e^{z_k}}{\sum_{j} e^{z_j}}\quad \text{and}\quad \phi(z)_c = \begin{cases}
1,\quad z = c\\
0,\quad z \ne c
\end{cases}
\end{align}
Our UFL loss is only applied to the target domain samples during training; source instances are trained with normal cross-entropy loss with label smoothing.
Additionally, we consider another use for $\Tilde{p}$: As written above, $\Tilde{p}$ contains information about how certain the model is for a class assignment. A peak of 100\% for one class reflects total certainty while a uniform distribution reflects total uncertainty. In the latter case, the sample should not be used for training as its pseudo-label is inaccurate with high probability. Based on this observation, we also propose Uncertainty-Based Filtering (UBF), which removes a sample from training if $\sum \text{top}_k(\Tilde{p}^{(i)}) \le \varphi$ where $\varphi$ is a fixed threshold. Parameter $k$ is dynamically set as $\frac{\lvert \mathcal{C} \lvert}{4}$ thus depending on the total number of classes in a dataset and considering the rising uncertainty that comes with more classes. Intuitively, this enforces the top classes to contain the majority of the probability mass. The UBF process also removes samples from the UFM calculation, therefore providing a cleaner estimate for the class means.

\section{Experiments}

\begin{figure}[t!]
\vspace{0.5cm}
\centering
\begin{tabular}{ccc}
\centered{\bmvaHangBox{
\includegraphics[width=1.5cm]{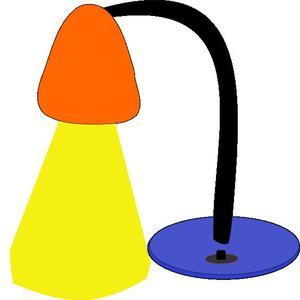}
\includegraphics[width=1.5cm]{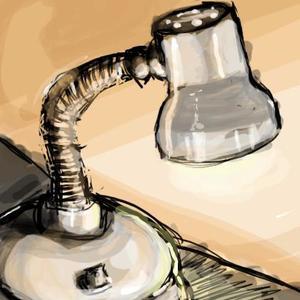}\\
\includegraphics[width=1.5cm]{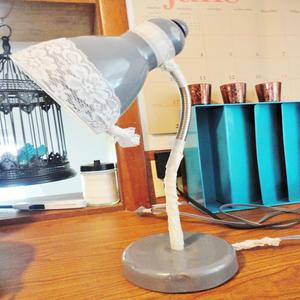}
\includegraphics[width=1.5cm]{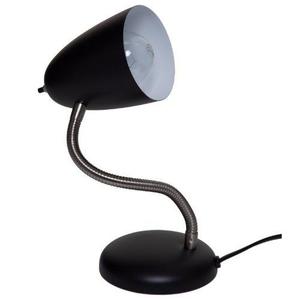}
}}&
\centered{\bmvaHangBox{
\includegraphics[width=1.5cm]{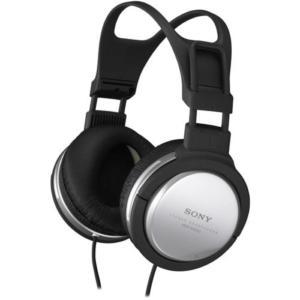}
\includegraphics[width=1.5cm]{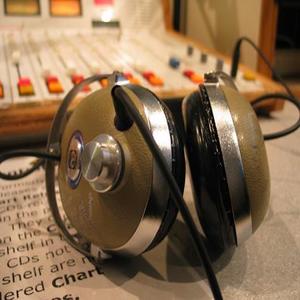}\\
\includegraphics[width=1.5cm]{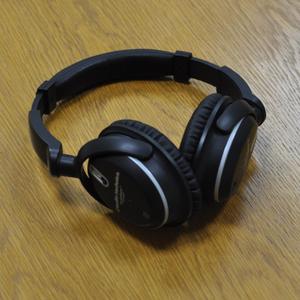}
\includegraphics[width=1.5cm]{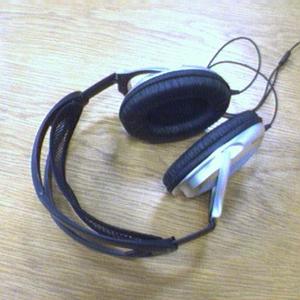}
}}&
\centered{\bmvaHangBox{
\includegraphics[width=1.5cm]{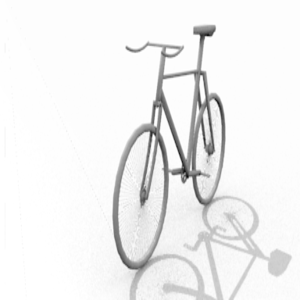}
\includegraphics[width=1.5cm]{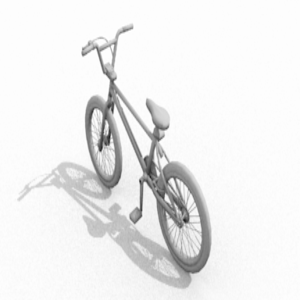}\\
\includegraphics[width=1.5cm]{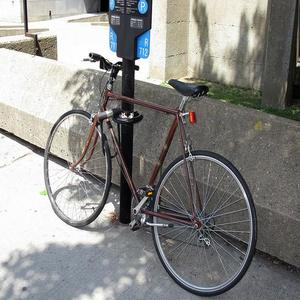}
\includegraphics[width=1.5cm]{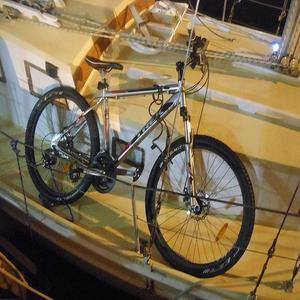}
}}\\
(a) Office-Home & (b) Office-Caltech & (c) VisDA 2017
\end{tabular}
\caption{Example images from the three datasets used for evaluation: (a) Office-Home with Art, Clipart, Product and Real-world domains. (b) Office-Caltech with Amazon, Caltech, DSLR and Webcam domains. (c) VisDA 2017 with synthetic and real domains.}
\label{fig:datasets}
\end{figure}
\subsection{Setup}

\textbf{Datasets}. We evaluate our proposed method on three public benchmark datasets: \textit{Office-Home}~\cite{officehome} is a challenging dataset with 15,588 images from 65 classes in the four domains Art, Clipart, Real-World and Product. Especially the Art and Clipart domains constitute a large domain gap to real-world data.
Additionally, we use the \textit{Office-Caltech} dataset with 2,533 images from 10 classes in the four domains Amazon, Caltech, DSLR and Webcam. 
Finally, we also evaluate on the Syn2Real-C (VisDA 2017) dataset~\cite{visda2017} with 152,397 synthetic 3D renders as well as 55,388 (validation set) and 72,372 (test set) real-world images from 12 classes. Examples from the three datasets are shown in Figure~\ref{fig:datasets}.\newline
\textbf{Hyperparameters}. For our Office-Home and Office-Caltech experiments, we use ResNet-50~\cite{he2016deep} as feature extractor; for VisDA 2017 ResNet-101 is the common setup. In both cases, networks are pretrained on ImageNet. We append one linear layer for classification and jointly optimize all parameters using SGD with Nesterov momentum~\cite{nesterov} of 0.95. 
When generating $\Tilde{p}$, we use $N_{{MC}}$=20 iterations with MC dropout rate 85\%. Features and $\hat{p}$ are calculated every 50 forward passes, $\overline{p}$ (and thus UFM) is resampled every five steps. For the purpose of SBL evaluation, we train on 4 NVIDIA GTX 1080 Ti GPUs.\newline
Our code is implemented in PyTorch~\cite{paszke2017automatic} and available on the project website\footnote{\url{https://gitlab.com/tringwald/ufal}}. Please refer to the supplementary material for further information about the training process.

\subsection{Results}

\begin{table}[t]
\vspace{0.25cm}
\centering
\resizebox{0.5\textwidth}{!}{
\begin{tabular}{lrr}
\toprule
    Method & Accuracy & Rel. Gain \\\hline
    Source only & 58.3 & +0.0\\
    BIS + source first & 63.0 & +4.7 \\
    BIS + random & 75.6 & +17.3\\
    BIS + SBL (random order) & 76.4 & +18.1 \\
    BIS + target first & 76.5 & +18.2\\
    BIS + SBL & 78.4 & +20.1\\
    BIS + SBL + UBF & 78.0 & +19.7\\
    BIS + SBL + UFL & 79.8 & +21.5\\
    BIS + SBL + UFL + UBF (no UFM) & 78.8 & +20.5\\
    BIS + SBL + UFL + UBF \textbf{(UFAL)} & \textbf{81.8} & \textbf{+23.5}\\
\toprule
\end{tabular}}\vspace{0.1cm}
\caption{\label{table:visda_ablation} Ablation study for the different parts of our proposed method on the VisDA 2017 validation set. Note that these results are reported as standard accuracy (in \%).}
\end{table}

\begin{table}[t]
\centering
\resizebox{\textwidth}{!}{
\begin{tabular}{clrrrrrrrrrrrrr}
\toprule
    Set & Method & {\scriptsize aero} & {\scriptsize bicycle} & {\scriptsize bus} & {\scriptsize car} & {\scriptsize horse} & {\scriptsize knife} & {\scriptsize motor} & {\scriptsize person} & {\scriptsize plant} & {\scriptsize skate} & {\scriptsize train} & {\scriptsize truck} & {\scriptsize Avg.}\\\hline
    Val & Source only & 64.7 & 30.8 & 70.1 & 69.8 & 80.1 & 20.2 & 85.4 & 25.0 & 73.1 & 36.7 & 85.9 & 7.1 & 54.1 \\
    Val & SimNet-152 \cite{simnet} & 94.3 & 82.3 & 73.5 & 47.2 & 87.9 & 49.2 & 75.1 & 79.7 & 85.3 & 68.5 & 81.1 & 50.3 & 72.9 \\
    Val & ADR \cite{adr} & 87.8 & 79.5 & 83.7 & 65.3 & 92.3 & 61.8 & 88.9 & 73.2 & 87.8 & 60.0 & 85.5 & 32.3 & 74.8 \\
    Val & SAFN \cite{xu2019larger} & 93.6 & 61.3 & 84.1 & 70.6 & 94.1 & 79.0 & \textbf{91.8} & 79.6 & 89.9 & 55.6 & \textbf{89.0} & 24.4 & 76.1 \\
    Val & DSBN \cite{dsbn} & 94.7 & 86.7 & 76.0 & 72.0 & 95.2 & 75.1 & 87.9 & 81.3 & 91.1 & 68.9 & 88.3 & 45.5 & 80.2 \\
    Val & DTA \cite{lee2019drop} & 93.7 & 82.8 & 85.6 & \textbf{83.8} & 93.0 & 81.0 & 90.7 & \textbf{82.1} & 95.1 & 78.1 & 86.4 & 32.1 & 81.5 \\
    Val & SE-152 \cite{french2017self} & 95.9 & \textbf{87.4} & 85.2 & 58.6 & 96.2 & 95.7 & 90.6 & 80.0 & 94.8 & 90.8 & 88.4 & 47.9 & 84.3 \\
    Val & CAN \cite{can} & 97.0 & 87.2 & 82.5 & 74.3 & \textbf{97.8} & \textbf{96.2} & 90.8 & 80.7 & \textbf{96.6} & \textbf{96.3} & 87.5 & \textbf{59.9} & \textbf{87.2} \\
    Val & \textbf{UFAL (ours)} & \textbf{97.6} & 82.4 & \textbf{86.6} & 67.3 & 95.4 & 90.5 & 89.5 & 82.0 & 95.1 & 88.5 & 86.9 & 54.0 & 84.7 \\\hline
    Test & CAN \cite{can} & --- & --- & --- & --- & --- & --- & --- & --- & --- & --- & --- & --- & 87.4 \\
    Test & SDAN~\cite{sdan} & 94.3 & 86.5 & 86.9 & 95.1 & 91.1 & 90.0 & 82.1 & 77.9 & 96.4 & 77.2 & 86.6 & 88.0 & 87.7\\ 
    Test & \textbf{UFAL (ours)} & 94.9 & 87.0 & 87.0 & 96.5 & 91.8 & 95.1 & 76.8 & 78.9 & 96.5 & 80.7 & 93.6 & 86.5 & \textbf{88.8}\\

\toprule
\end{tabular}}\vspace{0.cm}
\caption{\label{table:visda_sota}Classification accuracy (in \%) for different methods on the VisDA 2017 dataset. ResNet-101 is used as a backbone if not denoted by a hyphenated suffix.}
\end{table}

\textbf{Ablation Study}. We start by conducting an ablation study for every part of our proposed method and show the results on the VisDA 2017 dataset in Table~\ref{table:visda_ablation}. Note that these results are reported as standard accuracy instead of mean accuracy so that the class imbalance does not influence the evaluation. Our baseline is a model trained only on the source domain data. When adding our Binned Instance Sampling, this lower bound can already be improved by up to 20.1\%. However, we confirm that the multi-GPU batch-normalization (BN) is indeed highly dependent on the batch order (see section~\ref{sec:batchnorm}): Keeping only source data on the first replica leads to a degraded performance. This is expected as updates to $\mu$ and $\sigma$ of the BN layers are calculated purely over the source data, while the evaluation is on target data with a highly different distribution. Moving all target data to the front of the batch solves this and results in more than 13\% improvement compared to the previous setup. A random batch order is surprisingly competitive but worse than the target first setup. This is because a random order does not avoid extreme constellations such as only source data on a replica. We finally evaluate our proposed smart batch layout (SBL): Compared to other batch layouts, SBL achieves the highest accuracy of 78.4\%. We also show that it is necessary to keep samples from the same class together: The SBL (random order) setup is similar to SBL in terms of the 50\% split of source/target data per replica but uses randomly drawn source or target examples. This control experiment shows similar performance as the target first setup and reinforces the need for the SBL setup as proposed in section~\ref{sec:batchnorm}.\newline
Based on the BIS+SBL setup, we now also show the effectiveness of our proposed uncertainty-based loss and filtering in Table~\ref{table:visda_ablation}. Simply adding the UFL loss already results in a 1.4\% improvement. 
To control for the standalone effect of UBF, we also evaluate SBL+UBF, which shows no significant improvement over SBL alone. However, combining UFL and UFB into the Uncertainty-based Filtering and Feature Alignment (UFAL) leads to another 2\% improvement on top, beating all other evaluated setups. Additionally, we also control for the effect of the uncertain feature means (UFM) by not resampling from $\overline{p}$ but instead using $\hat{p}$ for class assignments. This decreases the accuracy and shows the need for UFM -- static feature means just encourage overfitting. Overall, our proposed UFAL method reaches the best accuracy of 81.8\%, improving the baseline by 23.5\%.

\textbf{Comparison to SOTA}. We now compare our proposed method to recent state-of-the-art approaches. Results for the VisDA 2017 datasets are shown in Table~\ref{table:visda_sota} and reported as average class accuracy, in accordance with the VisDA challenge evaluation metric. The test set labels were private up until recently due to being part of a challenge -- prior research thus focused on the validation set. On this subset, UFAL surpasses recent methods such as SimNet~\cite{simnet}, SAFN~\cite{xu2019larger}, DSBN~\cite{dsbn} and even SE~\cite{french2017self} -- the winner of the VisDA 2017 challenge. While CAN~\cite{can} is slightly better on the validation set, this does not transfer to the real test set where UFAL leads with 88.8\%. On the test set, UFAL also outperforms SDAN~\cite{sdan}, which uses an ensemble of four different network architectures and multiple runs for domain adaptation. Given the current VisDA 2017 leaderboard, UFAL would rank 2\textsuperscript{nd} place -- only behind a 5$\times$ ResNet-152 ensemble with results averaged over 16 test time augmentation runs. However, this is not a fair comparison to our single ResNet-101 setup.

Additionally, Table~\ref{table:officecaltech_comparison} compares UFAL to SOTA methods on the Office-Caltech dataset. Again, we surpass recently proposed methods such as GTDA+LR~\cite{vascon2019unsupervised} and RWA~\cite{rwa}. UFAL even outperforms the ensemble based algorithm of Rakshit~\etal~\cite{ccduda} by 0.5\%.

Finally, we also report results on the challenging Office-Home dataset in Table~\ref{table:officehome_comparison}. Yet again, our proposed UFAL approach outperforms recent SOTA methods such as CADA-P~\cite{kurmi2019attending}, TADA~\cite{wang2019transferable} and SymNets~\cite{zhang2019domain}.
Overall, our experiments show that UFAL can achieve state-of-the-art results on a wide variety of tasks and perform unsupervised domain adaptation even in complex setups such as learning from synthetic, product or clipart images.

\begin{table}[t]
\centering
\resizebox{1.0\textwidth}{!}{
\begin{tabular}{@{\extracolsep{4pt}}lrrrrrrrrrrrrr@{}}
\toprule
    \multirow{2}{*}{Method} & \multicolumn{3}{c}{A} & \multicolumn{3}{c}{C} & \multicolumn{3}{c}{D} & \multicolumn{3}{c}{W} & \multirow{2}{*}{Avg.}\\\cline{2-4}\cline{5-7}\cline{8-10}\cline{11-13}
     & C & D & W & A & D & W & A & C & W & A & C & D & \\\hline
    RTN \cite{rtn} & 88.1  & 95.5  & 95.2  & 93.7  & 94.2  & 96.9  & 93.8  & 84.6  & 99.2  & 92.5  & 86.6  & \textbf{100.0}  & 93.4 \\
    Rahman et al. \cite{rahman2019} & 89.1  & 96.6  & 95.7  & 93.6  & 93.4  & 95.2  & 94.7  & 84.7  & 99.4  & 94.8  & 86.5  & \textbf{100.0}  & 93.6  \\
    ADACT \cite{li2019adversarial} & 92.7  & 96.5  & 95.0  & 94.3  & 93.0  & 93.7  & 94.9  & 88.7  & 99.0  & 94.2  & 90.3  & \textbf{100.0}  & 94.4  \\
    GTDA+LR \cite{vascon2019unsupervised}  & 91.5 & 98.7 & 94.2 & 95.4 & 98.7 & 89.8 & 95.2 & 89.0 & 99.3 & 95.2 & 90.4 & \textbf{100.0} & 94.8\\
    RWA \cite{rwa} & 93.8 & 98.9 & 97.8 & 95.3 & \textbf{99.4} & 95.9 & 95.8 & 93.1 & 98.4 & 95.3 & 92.4 & 99.2 & 96.3\\
    Rakshit et al. \cite{ccduda} & 92.8 & 98.9 & 97.0 & \textbf{96.0} & 99.0 & 97.0  & \textbf{96.5}  & \textbf{97.0}  & 99.5  & 95.5  & 91.5  & \textbf{100.0}  & 96.8\\
    \textbf{UFAL (ours)} & \textbf{95.1} & \textbf{99.4} & \textbf{99.7} & \textbf{96.0} & 96.8 & \textbf{99.7} & 95.8 & 95.0 & \textbf{99.7} & \textbf{96.3} & \textbf{95.0} & 99.4 & \textbf{97.3} \\
\toprule
\end{tabular}}\vspace{0.01cm}
\caption{\label{table:officecaltech_comparison}Classification accuracy (in \%) for different methods on the Office-Caltech dataset with domains Amazon, Caltech, DSLR and Webcam.}
\end{table}

\begin{table}[t]
\centering
\resizebox{1.0\textwidth}{!}{
\begin{tabular}{@{\extracolsep{4pt}}lrrrrrrrrrrrrr@{}}
\toprule
    \multirow{2}{*}{Method} & \multicolumn{3}{c}{Ar} & \multicolumn{3}{c}{Cl} & \multicolumn{3}{c}{Pr} & \multicolumn{3}{c}{Rw} & \multirow{2}{*}{Avg.}\\\cline{2-4}\cline{5-7}\cline{8-10}\cline{11-13}
     & Cl & Pr & Rw & Ar & Pr & Rw & Ar & Cl & Rw & Ar & Cl & Pr & \\\hline
    CDAN+E \cite{long2018conditional} & 50.7 & 70.6 & 76.0 & 57.6 & 70.0 & 70.0 & 57.4 & 50.9 & 77.3 & 70.9 & 56.7 & 81.6 & 65.8\\
    MDDA \cite{wang2020transfer} & 54.9 & 75.9 & 77.2 & 58.1 & 73.3 & 71.5 & 59.0 & 52.6 & 77.8 & 67.9 & 57.6 & 81.8 & 67.3\\
    TADA \cite{wang2019transferable} & 53.1	& 72.3 & 77.2 & 59.1 & 71.2 & 72.1 & 59.7 & 53.1 & 78.4 & 72.4 & 60 & 82.9 & 67.6 \\
    SymNets \cite{zhang2019domain} & 47.7 & 72.9 & 78.5 & 64.2 & 71.3 & 74.2 & 64.2 & 48.8 & 79.5 & 74.5 & 52.6 & 82.7 & 67.6\\
    CDAN+TransNorm \cite{wang2019transferableNIPS} & 50.2 & 71.4 & 77.4 & 59.3 & 72.7 & 73.1 & 61.0 & 53.1 & 79.5 & 71.9 & 59.0 & 82.9 & 67.6\\
    CADA-P \cite{kurmi2019attending} & 56.9 & \textbf{76.4} & \textbf{80.7} & 61.3 & \textbf{75.2} & \textbf{75.2} & 63.2 & 54.5 & \textbf{80.7} & \textbf{73.9} & 61.5 & \textbf{84.1} & 70.2\\
    \textbf{UFAL (ours)} & \textbf{58.5} & 75.4 & 77.8 & \textbf{65.2} & 74.7 & 75.0 & \textbf{64.9} & \textbf{58.0} & 79.9 & 71.6 & \textbf{62.3} & 81.0 & \textbf{70.4} \\
\toprule
\end{tabular}}
\caption{\label{table:officehome_comparison}Classification accuracy (in \%) for different methods on the Office-Home dataset with domains Art, Clipart, Product and Real-world.}
\end{table}

\begin{figure}[t!]
\vspace{0.25cm}
\centering
\begin{tabular}{cc}
\centered{\bmvaHangBox{
\includegraphics[width=0.55\textwidth]{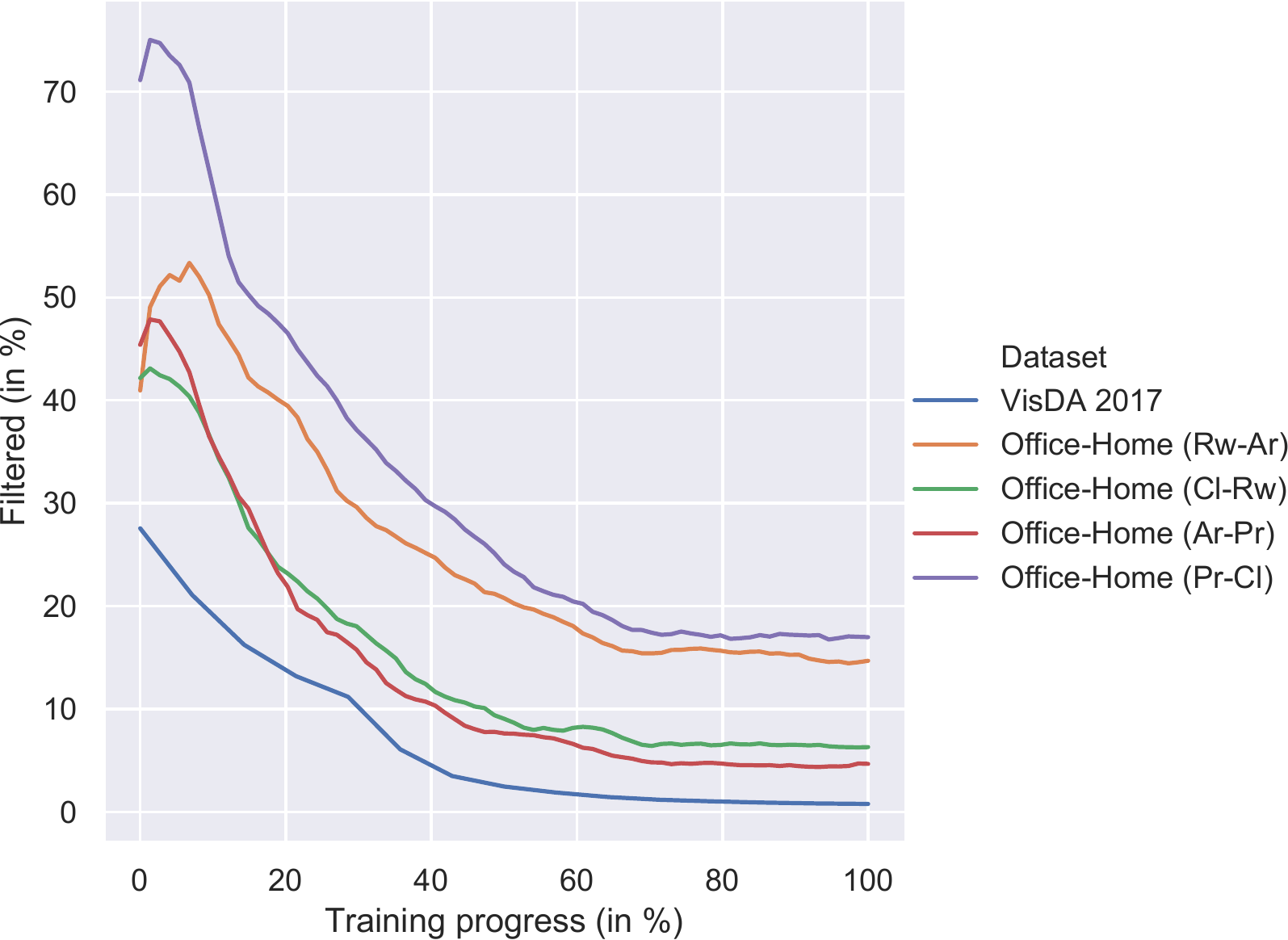}
}}&
\centered{\bmvaHangBox{
\includegraphics[width=0.3\textwidth]{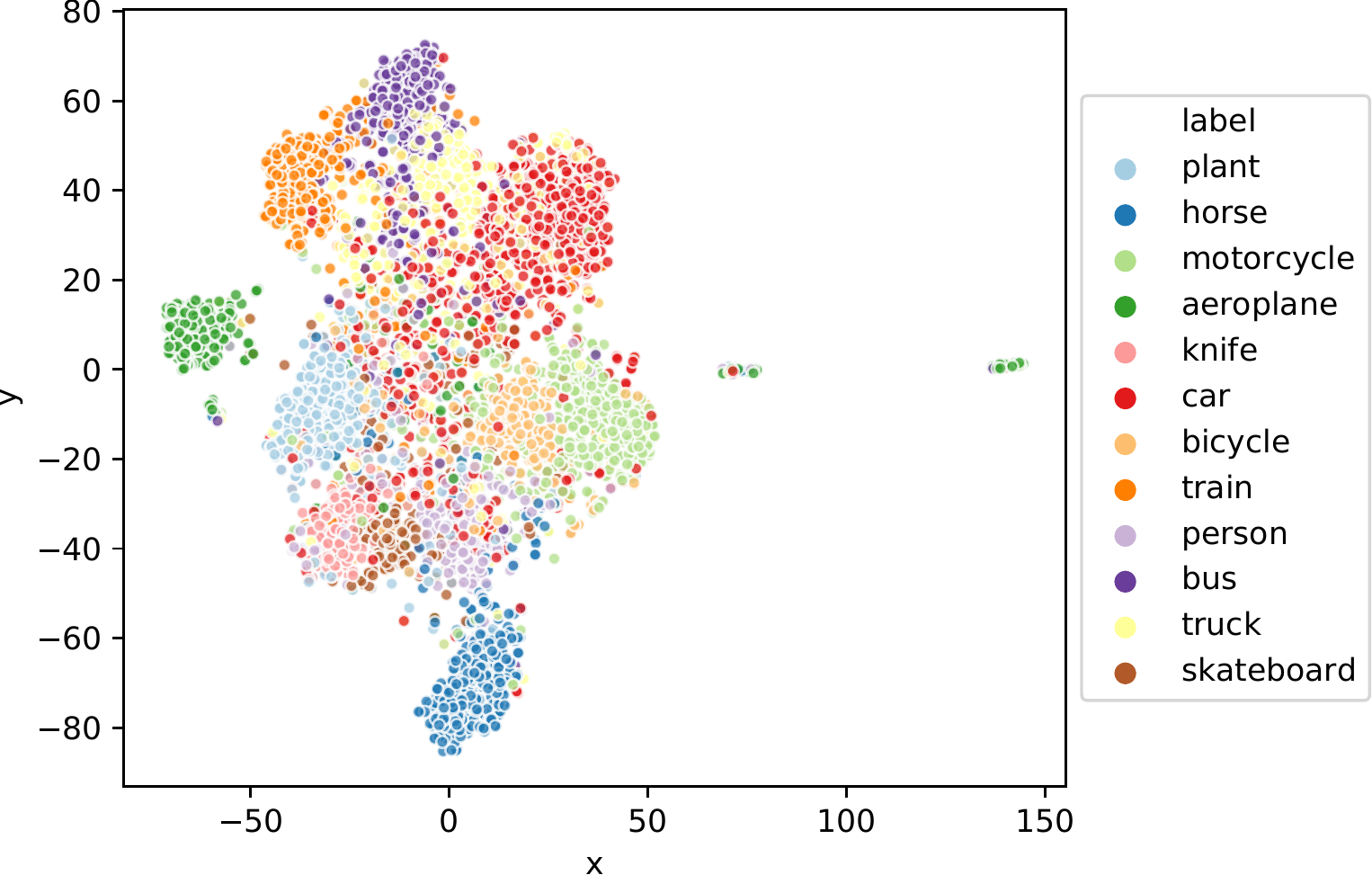}\\
\includegraphics[width=0.3\textwidth]{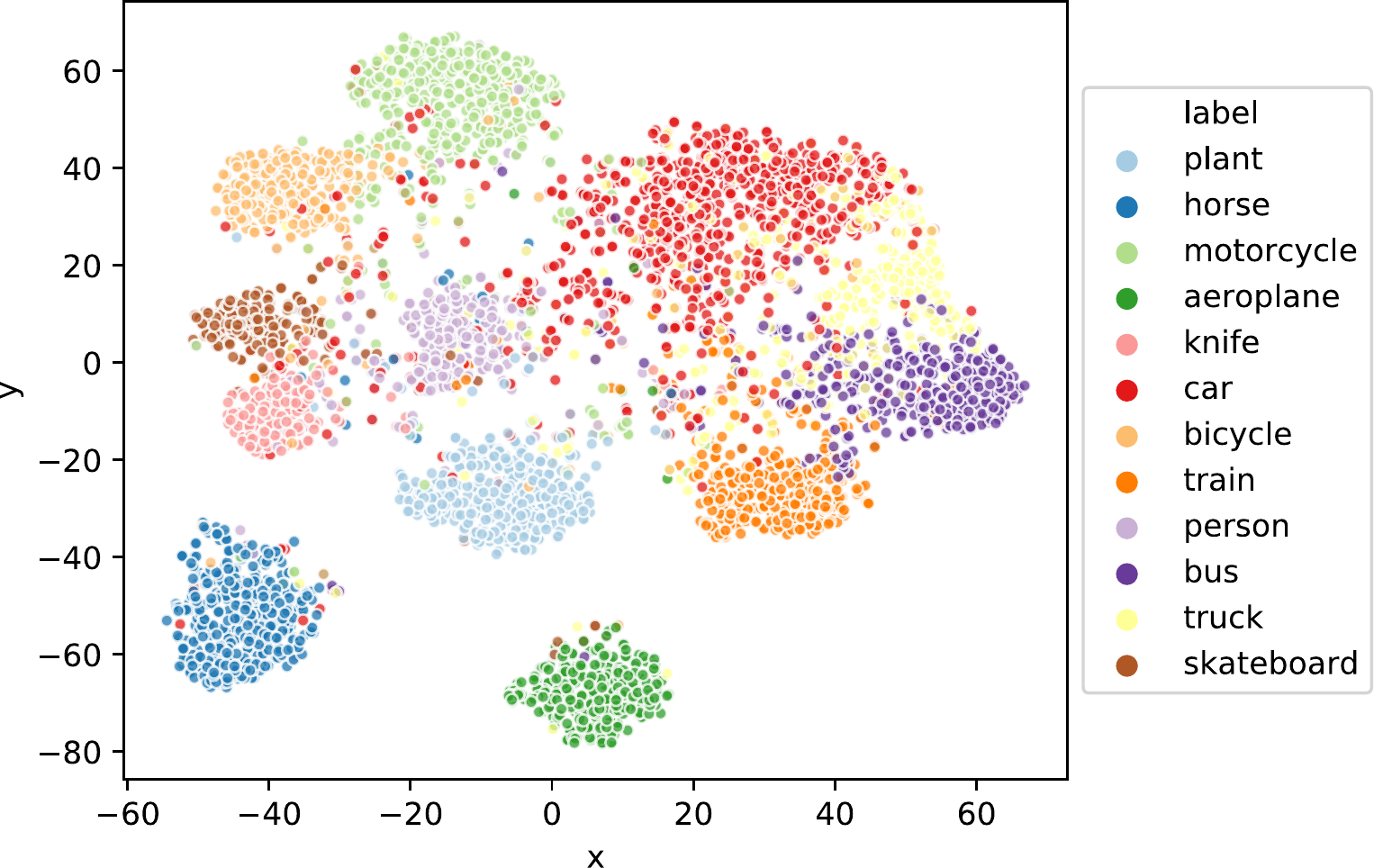}
}}\\
(a) Uncertainty-based Filtering & (b) Feature visualization
\end{tabular}
\caption{(a): Percentage of target domain samples filtered by the UBF process as the adaptation phase progresses. (b): t-SNE visualization of features extracted from the VisDA 2017 target domain before the adaptation process (top) and after the adaptation process (bottom). Best viewed in the digital version.}
\label{fig:results_visualization}
\end{figure}

\textbf{Visualizations}. In Figure~\ref{fig:results_visualization}a, we show the number of target domain samples filtered by UBF as training progresses. Expectedly, the amount of filtered samples starts at a high percentage due to the initial uncertainty that eventually declines as training converges. We note that the number of filtered samples that remain at the end of the adaptation phase correlates with the final accuracy of the considered datasets as well as their number of classes. This is also expected, because the difficulty of the transfer task and the number of classes increase the uncertainty of any given prediction and thus the filtering process. For the VisDA 2017 dataset, the filtered percentage drops to almost 0\% due to its distinguishable 12 classes, while the filtered percentage remains at \textasciitilde 17\% for Office-Home's hardest transfer task (Pr-Cl) with 65 classes.

Furthermore, we provide t-SNE visualizations for the VisDA 2017 validation set features in Figure~\ref{fig:results_visualization}b. Evidently, the model has not learned discriminative target representations after the source domain training: Almost all features are densely packed into a single cluster. After the adaptation process with UFAL, the features are separated into clearly distinct clusters. Due to UFAL's alignment properties, related classes are also kept close in feature space (\eg car, truck and bus), therefore also providing a semantical interpretation. Further discussion on these topics is provided in the supplementary material.

\section{Conclusion}

In this paper, we explore the usage of a model's predictive uncertainty for unsupervised domain adaptation. Our proposed Uncertainty-based Filtering and Feature Alignment (UFAL) method exploits uncertainty for filtering of training data and alignment of features in Euclidean space. Additionally, we extend the concept of ghost batch-normalization to UDA tasks and uncover the importance of a smart batch layout for multi-GPU training.
We evaluate UFAL's efficacy on three commonly used UDA benchmark datasets and achieve state-of-the art results even when compared to strong baselines. Our code will be made available to the community for reproduction of our results and to encourage further research.

\clearpage
\bibliography{egbib}
\end{document}